\documentclass{article}

\usepackage[preprint]{neurips_2026}

\usepackage[utf8]{inputenc}
\usepackage[T1]{fontenc}
\usepackage{amsfonts}
\usepackage{amsmath, amssymb}
\usepackage{txfonts}
\usepackage[hidelinks]{hyperref}
\usepackage{url}
\usepackage{booktabs}
\usepackage{nicefrac}
\usepackage{microtype}
\usepackage{xcolor}
\usepackage{multirow}
\usepackage{graphicx}
\graphicspath{{figures/}}

\title{The Translation Tax Is Not a Scalar: A Counterfactual Audit of English-Source Cue Inheritance in Chinese Multilingual Benchmarks}

\author{%
Zezheng Lin\thanks{Correspondence to Zezheng Lin.} \And Fengming Liu \And Handi Li }

\begin{document}
\maketitle

\begin{abstract}
The Translation Tax is often treated as a scalar: translated benchmarks are assumed to inflate scores by preserving English-source cues. We audit this claim in an English-to-Chinese setting. Three proxy estimators disagree: back-translation gaps are small and parser-fragile; cue-score calibration does not predict item-level gains; and a six-model native-control comparison shows model-family rather than uniform benchmark effects. We add a same-item LLM-naturalization stress test that holds answer, options, and content fixed while rewriting Chinese surface form. After correcting a prompt-construction bug, this contrast no longer supports a model-family interaction, but it preserves a residue dose-response: high-residue items benefit while low-residue items do not. The result is not a single Translation Tax, but a set of estimator- and item-dependent validity risks. We release per-cell evidence, the naturalization protocol, human QC, and a reporting checklist for translated multilingual benchmark papers.
\end{abstract}

\section{Introduction}

The de facto standard for evaluating multilingual capability of LLMs depends on translated benchmarks. MMMLU is the multilingual translation of MMLU \citep{hendrycks2021mmlu} released by OpenAI in 2024; the released MMMLU test split used here contains $14{,}042$ items across $57$ subjects, and the Chinese (\texttt{ZH\_CN}) subset matches this count.\footnote{We confirmed the released item count programmatically against the OpenAI MMMLU dataset card; earlier drafts of this paper reported $15{,}908$ items following a citation that conflated the MMLU benchmark size with a related dataset.} Belebele \citep{bandarkar2024belebele} is a parallel reading comprehension dataset spanning 122 language variants with approximately 900 items per variant. XCOPA \citep{ponti2020xcopa} covers 11 languages with all non-English data translated from English COPA. The translation processes were carefully implemented, but the structural fact that all non-English items derive from a common English source introduces a systematically underestimated side effect: model performance on non-English items may reflect not only true language understanding but also identification of English cues preserved through translation. \citet{artetxe2020translation} demonstrated precisely this mechanism in cross-lingual transfer, showing that translation introduces subtle artifacts that models exploit as non-semantic shortcuts.

The NeurIPS 2025 workshop \emph{Centering Low Resource Languages and Cultures in the Age of Large Language Models} demonstrated rapidly rising community attention to multilingual evaluation. \citet{sitaram2025} explicitly identified three core challenges of multilingual benchmark evaluation: coverage, representativeness, and trust/scientific rigor. Translated English benchmarks were listed alongside US-centric framing as the two main representativeness deficiencies. \citet{wu2025} sharpened this concern by showing that translated benchmarks align far worse with local human judgments than natively constructed alternatives (Spearman $0.47$ vs.\ $0.68$). Yet rigorous quantification of how much benchmark score is driven by translation cues rather than true language understanding remains scarce.

\paragraph{Contributions.} The paper makes three contributions.
\begin{enumerate}
\item \textbf{Conceptual.} The Translation Tax is defined as the cue-driven
component of translation-induced score shift, distinguished from semantic
translation error. Each estimator is paired with explicit identification
assumptions and named failure modes; a fourth estimator (E4, matched
LLM-naturalization stress test) is introduced to hold item content fixed
between conditions.
\item \textbf{Empirical.} Three proxy estimators on a 228-item MMMLU and
100-item Belebele subset converge on the same diagnosis: small
back-translation gaps that are parser-fragile, an annotation calibration that
does not support item-level cue exploitation, and a six-model native-control
comparison in which the largest gaps appear within the Chinese-optimized
subgroup rather than the English-centric one. E4 (after a corrected
prompt-construction bug) returns a small positive average effect concentrated
on items selected ex ante for high translation residue
($\Delta_{\text{high}}=+0.103$ vs.\ $\Delta_{\text{low}}=-0.015$ excl.\
parser outlier) and no statistically significant model-family interaction.
The Translation Tax is not a single scalar correction but an estimator- and
item-dependent set of validity risks.
\item \textbf{Reporting.} A translation-cue identifiability reporting
checklist covers estimator scope, parser fragility, and subgroup contrasts
that translated multilingual benchmark papers can adopt as a standard
validity dimension at submission time.
\end{enumerate}

\paragraph{Scope.} Point estimates are small ($1$--$5$ percentage points) and most
individual cells do not exclude zero. The paper reports proxy estimates with their identification assumptions disclosed; the latent Translation Tax is not directly measured by any single estimator. The annotation estimator is executed only as a 30-item single-annotator rubric calibration, treated as calibration evidence rather than a full bilingual-human estimator (Section~\ref{sec:annotation_pilot}). The study uses English-to-Chinese as a single case pair to enable detailed within-pair analysis rather than thin breadth across many languages.

\section{Related Work}

XNLI \citep{conneau2018xnli} extending MultiNLI to 14 non-English languages explicitly discussed translation effects on evaluation validity. XCOPA \citep{ponti2020xcopa} discussed translation effects in its Limitations section. Belebele \citep{bandarkar2024belebele} detailed its FLORES-200 translation pipeline quality control while acknowledging traces detectable by models. \citet{artetxe2020translation} systematically analyzed how translation artifacts in cross-lingual benchmarks create exploitable shortcuts: independent translation of premises and hypotheses reduced lexical overlap relative to English originals, introducing a spurious signal. Their work is the most direct predecessor to ours, though they focused on diagnosing artifact types rather than quantifying score inflation across benchmarks.

\citet{wu2025}, surveying over 2{,}000 multilingual benchmarks published between 2021 and 2024, provided the most comprehensive recent evidence that translation quality directly undermines the validity of human preference alignment in multilingual evaluation. \citet{clark2020tydiqa} introduced TyDi QA, designed to avoid translationese artifacts by collecting questions natively. INCLUDE \citep{romanou2025include} sources items natively from regional exam pools across 44 languages; we use its Chinese subset as the non-translation control source for E3.

Prompt sensitivity \citep{mizrahi2024multiprompt, zhuo2024prosa} is complementary to our focus on the linguistic validity of evaluation data. The work nearest to ours is \citet{artetxe2020translation}, which we extend by separating cue inheritance from translation noise and by quantifying score-level effects rather than diagnosing artifact types.

\section{Formal Definitions and Identification}
\label{sec:formal}

\begin{figure}[h]
\centering
\includegraphics[width=0.92\linewidth]{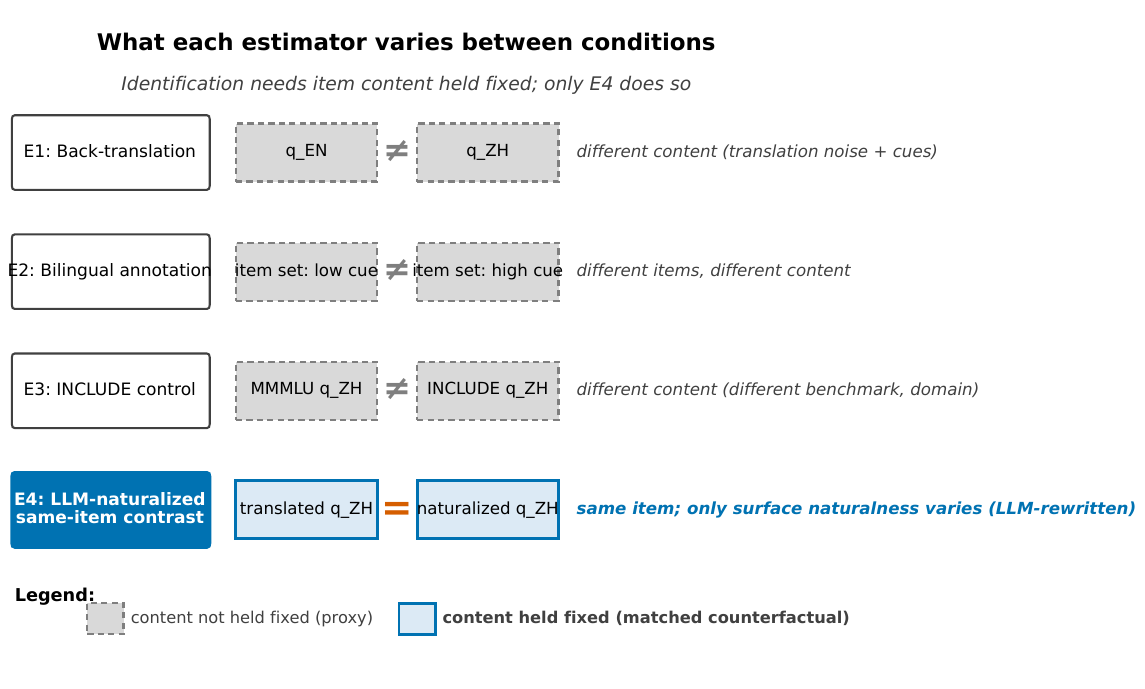}
\caption{What each estimator varies between conditions. E1, E2, and E3 each
contrast scores across items that differ in content (translation noise, item
identity, or benchmark register), so none of them holds content fixed. E4
(matched naturalization, Section~\ref{sec:e4_naturalization}) is the only
contrast in the paper that varies surface naturalness with content held fixed.}
\label{fig:triangulation}
\end{figure}

\paragraph{Translation Tax.} For a translated benchmark $B$ in target language $L$ and model
$M$, let $\text{score}_L(M)$ denote the observed score on the translated benchmark and $\text{score}_L^{\text{natural}}(M)$ denote the (unobservable) score on a hypothetical natural-$L$ version of the same items. The translation effect decomposes as: \[ \text{score}_L(M) - \text{score}_L^{\text{natural}}(M) = \text{TT} + E, \] where $\text{TT}$ is the Translation Tax (cue-driven inflation from residual source-language structures) and $E$ is the Semantic Error Effect (degradation from meaning distortion). Since $\text{score}_L^{\text{natural}}$ is unobservable, TT is not directly identified from a single observable contrast. We use three complementary proxy estimators, each with its own identification assumptions, and triangulate (Figure~\ref{fig:triangulation}).

\subsection{Estimator E1 (Back-Translation): Identification and Failure Modes}

For an English-centric model $M_{\text{EN}}$, let $T_{\text{back}}$ denote the L-to-English back-translation pipeline and $q_L^{\text{back}} = T_{\text{back}}(q_L)$. The back-translation estimator is: \[ \text{TT}_{\text{back}}(B, L, M_{\text{EN}}) = \text{score}_{\text{EN}}^{q_{\text{EN}}}(M_{\text{EN}}) - \text{score}_{\text{EN}}^{q_L^{\text{back}}}(M_{\text{EN}}). \] TT$_{\text{back}}$ measures the score gap that an English-centric model exhibits between original-English items and items round-tripped through L. If back-translation preserves semantic content, this gap reflects the structural cue residue lost in the round trip.

\paragraph{Failure modes (where E1 does not identify TT).} (F1) \emph{Back-translation
noise.} Real pipelines introduce semantic noise $E_{\text{back}}$, so $\text{TT}_{\text{back}} \approx \text{TT} + E_{\text{back}}$, making the estimator a noisy upper bound. BLEU/BERTScore QC reduces but does not eliminate $E_{\text{back}}$. (F2) \emph{Naturalness loss.} Back-translation may produce grammatical but unnatural English; the observed gap can include English-text-quality penalties unrelated to cue inheritance. (F3) \emph{Asymmetric translation difficulty.} Some content (e.g., Chinese-specific cultural items in MMMLU localized subjects) may be harder to back-translate than to translate forward, biasing $E_{\text{back}}$ in subject-correlated ways. The paper therefore does not interpret a positive $\text{TT}_{\text{back}}$ as evidence of cue exploitation by itself; it functions as one of three triangulating signals.

\subsection{Estimator E2 (Bilingual Native Annotation): Calibration Status}

For a stratified random sample of items, two or more bilingual native annotators score each item on a 5-point Likert scale on three dimensions---\emph{cue identifiability}, \emph{cultural residue}, and \emph{syntactic residue}---and Spearman correlation between cue identifiability and item-level TT$_{\text{back}}$ signal is computed by model subgroup. The full protocol blinds annotators to all model outputs, uses anchor-calibrated rubric prompts, includes at least one external bilingual annotator, and reports weighted Cohen's $\kappa$ and Krippendorff's $\alpha$ on the ordinal Likert scale (target weighted $\kappa > 0.7$).

\paragraph{What is executed in this study.} A 30-item single-annotator rubric-calibration
run with one LLM-rubric annotator. This is calibration evidence on rubric coverage and score distribution, not the bilingual-human estimator. It cannot support or refute the cue-exploitation hypothesis at this scale and is reported as a null calibration result (Section~\ref{sec:annotation_pilot}). Triangulation in the results section therefore rests on E1 and E3; E2 enters as a calibration check only.

\subsection{Estimator E3 (INCLUDE Non-Translation Control): Identification Scope}

INCLUDE \citep{romanou2025include} sources items natively from Chinese regional exam pools rather than via translation; INCLUDE-base-44 Chinese contains 545 four-choice test items spanning 57 topics. The paper compares aggregate (per-model) accuracy on translated benchmark Chinese against INCLUDE Chinese.

\paragraph{Identification scope.} INCLUDE Chinese items are drawn from different content
distributions than MMMLU Chinese (regional exams vs.\ academic subjects). The comparison is unmatched at the item level: it does not match items on difficulty, topic, or register, and is therefore an external-validity comparison rather than an item-level counterfactual. Item-level effects cannot be isolated from content-distribution effects under this design.

\section{Experimental Design}
\label{sec:design}

\subsection{Benchmarks and Models}

\paragraph{Benchmarks.} (1) MMMLU Chinese subset: the \texttt{ZH\_CN} test split released
by OpenAI contains 14{,}042 items across 57 subjects, matching the MMLU test split used by the released MMMLU dataset; we draw a stratified subset of 228 items at 4 per subject (seed 42). (2) Belebele Chinese subset (\texttt{zho\_Hans}; 900 items); subset of 100 items (seed 42). (3) INCLUDE Chinese subset as non-translation control (545 test items, 4-choice).

\paragraph{Models.} Nine total in three groups, all accessed via a unified provider gateway:
\begin{itemize}
\item Group A (English-centric frontier): gpt-4o, gpt-4o-mini, gpt-5.4-mini.
\item Group B (Chinese-optimized): deepseek-chat, qwen-max, glm-4.5.
\item Group C (Open-source multilingual mid-sized): llama-3.3-70b-instruct,
qwen2.5-72b-instruct, glm-4-air.
\end{itemize}
Anthropic Claude and Google Gemini could not be included because the gateway provider's channels for both were unavailable during the analysis window. This narrows Group A's coverage and is discussed in Limitations.

\subsection{Scoring Protocol}

\paragraph{Scoring protocol.} Provider-API access does not uniformly expose log-probabilities
across all nine models, so we use a \textbf{fixed-prompt, single-letter-extraction protocol}, not the \texttt{lm-evaluation-harness} library. Specifically: zero-shot prompting with the question, four labeled choices, and an instruction to respond with a single letter (A, B, C, or D); answer parsing extracts the first matching letter from the response. We report parser validity rates per model (the proportion of responses that yielded a parsable letter); models with low validity rates are flagged. We use 0-shot for all benchmarks at this stage; a 5-shot replication is not part of this study.

\paragraph{Background and rationale.} The original literature on these benchmarks reports 5-shot
results for MMMLU. The 0-shot setup here is therefore not directly comparable to published benchmark numbers in absolute terms; it is internally consistent across our three quantities ($q_{\text{ZH}}$, $q_{\text{EN}}$, $q_{\text{back}}$) for the same model, which is what TT$_{\text{back}}$ requires.

\paragraph{Validity statistics.} Across the MMMLU 228-item sample, eight of nine models achieve
parser validity $\geq 0.978$. The exception is gpt-5.4-mini ($0.825$), which sometimes returns multi-token reasoning before producing a letter; we report results for this model separately and exclude its $q_{\text{ZH}}$ items with parser failure from per-model accuracy estimates.

\subsection{Back-Translation Pipeline}

We use a commercial LLM (\texttt{deepseek-chat}) as the primary back-translation pipeline because (a) it produced higher BLEU scores than NLLB-200 distilled in preliminary tests, (b) the local NLLB-200 1.3B model exceeded our sandbox memory budget. Note that this conflicts with the methodological preference for an open reproducible pipeline; an NLLB-200 distilled-600M comparison is not part of this study.

\paragraph{Quality control.} For each item, we compute BLEU between $q_{\text{EN}}$ (the
original English item) and $q^{\text{back}}_L$ (the back-translated item). We attempted BERTScore F1 with \texttt{roberta-large}; the model download exceeded our sandbox disk budget during this analysis, so BLEU is the sole QC criterion. Items with BLEU $< 0.30$ are flagged \texttt{excluded\_qc} and strictly excluded from the TT$_{\text{back}}$ paired analysis.

\paragraph{QC pass rates.} MMMLU sample: 213/228 items pass ($93.4\%$, mean BLEU $0.568$).
Belebele sample: 87/100 items pass ($87.0\%$, mean BLEU $0.404$).

\paragraph{Source alignment audit.} A separate concern is whether the $q_{\text{EN}}$,
$q_{\text{ZH}}$, and $q^{\text{back}}$ records correspond to the same underlying source item. Misaligned English references would mechanically bias TT$_{\text{back}}$. We ran a triple-alignment audit against the MMMLU 228-item sample: all 228 IDs align across the three sources, and all 228 items have matching choice-counts (4 each). We then checked whether the gold-answer index is consistent between $q_{\text{EN}}$ and $q_{\text{ZH}}$ as a content-level alignment proof, and found 2 mismatches (items \texttt{05954}, \texttt{06001}, both in \texttt{high\_school\_world\_history}): the Chinese item and English item are entirely different questions, indicating a source-side alignment error in MMMLU \texttt{ZH\_CN} for those positions. Both flagged items are already excluded by the BLEU $< 0.30$ QC filter (BLEU $= 0.041$ and $0.025$), so the strict-QC subset below is unaffected. The audit script and detailed results are released as \texttt{run\_alignment\_audit.py} and \texttt{results/alignment\_audit.json}.

\section{Results}
\label{sec:results}

We report \textbf{strict-QC} TT$_{\text{back}}$ estimates: paired comparisons restricted to items where (a) the back-translation passed BLEU QC, (b) the model produced parsable answers on both $q_{\text{EN}}$ and $q^{\text{back}}_L$. This corrects an issue in the initial earlier scoring run where some QC-failed items were included.

\subsection{TT$_{\text{back}}$ Point Estimates and Confidence Intervals}

Table~\ref{tab:tt_back_strict} reports the strict-QC TT$_{\text{back}}$ paired bootstrap estimates (10{,}000 resamples, seed 42) for the three Group-A models on the two benchmarks.

\begin{table}[h]
\centering
\caption{Strict-QC TT$_{\text{back}}$ estimates (paired bootstrap, 10{,}000 resamples).
Strict QC restricts to items with BLEU $\geq 0.30$ \emph{and} valid model predictions on both
$q_{\text{EN}}$ and $q^{\text{back}}_L$. Discordant counts are paired items where one set is
correct and the other incorrect; ties are paired items where both are correct or both are
incorrect.}
\label{tab:tt_back_strict}
\small
\begin{tabular}{llrrrrrrr}
\toprule
\textbf{Bench} & \textbf{Model} & \textbf{n} & $q_{\text{EN}}$ & $q^{\text{back}}$ &
\textbf{TT$_{\text{back}}$} & \textbf{95\% CI} & \textbf{Pos/Neg} & \textbf{Sign p} \\
\midrule
\multirow{3}{*}{MMMLU}
 & gpt-4o       & 209 & 0.818 & 0.789 & $+0.029$ & $[-0.005,+0.067]$ & 10/4 & 0.180 \\
 & gpt-4o-mini  & 209 & 0.746 & 0.742 & $+0.005$ & $[-0.038,+0.048]$ & 11/10 & 1.000 \\
 & gpt-5.4-mini & 169 & 0.876 & 0.828 & $+0.047$ & $[+0.006,+0.095]$ & 12/4 & 0.077 \\
\midrule
\multirow{3}{*}{Belebele}
 & gpt-4o       & 87 & 0.977 & 0.954 & $+0.023$ & $[+0.000,+0.057]$ & 2/0 & 0.500 \\
 & gpt-4o-mini  & 87 & 0.954 & 0.954 & $+0.000$ & $[-0.046,+0.046]$ & 2/2 & 1.000 \\
 & gpt-5.4-mini & 82 & 0.976 & 0.963 & $+0.012$ & $[-0.024,+0.061]$ & 2/1 & 1.000 \\
\bottomrule
\end{tabular}
\end{table}

\begin{figure}[h]
\centering
\includegraphics[width=0.85\linewidth]{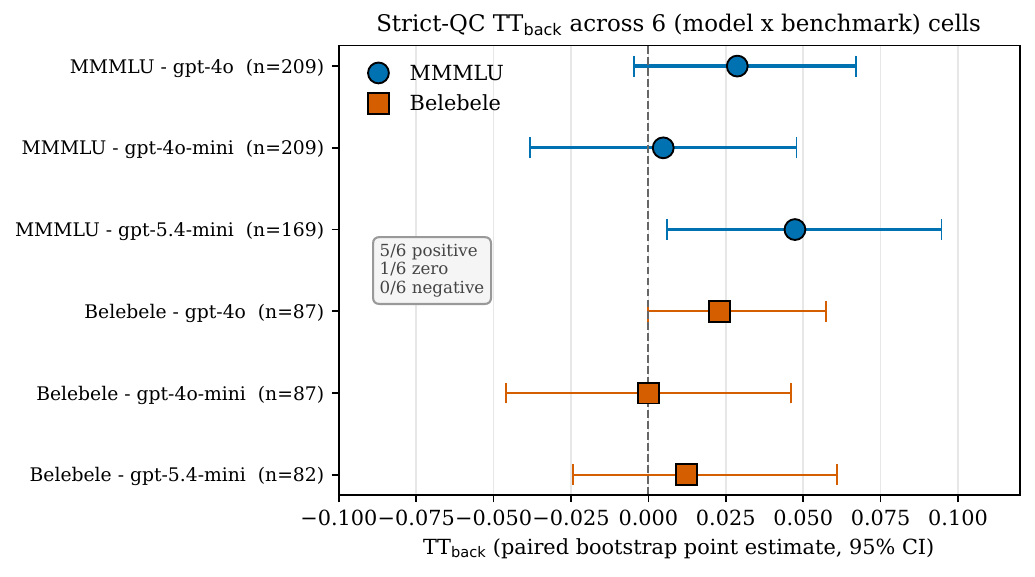}
\caption{Strict-QC TT$_{\text{back}}$ across six (model $\times$ benchmark) cells.
Circles: MMMLU; squares: Belebele. Point estimates with 95\% paired-bootstrap
CIs. Five of six cells are positive and one is exactly zero; most individual
CIs contain zero. Effect sizes are small in magnitude (range $0.000$--$0.047$).
The largest cell (MMMLU $\times$ gpt-5.4-mini) is parser-fragile.}
\label{fig:forest_tt_back}
\end{figure}

\paragraph{Empirical pattern.} Five of six cells produce positive point estimates and one is
exactly zero; no cell is negative. One cell, MMMLU/gpt-5.4-mini, has a bootstrap CI that narrowly excludes zero ($[+0.006, +0.095]$), but this is also the model with the lowest parser validity (Section~\ref{sec:design}) and does not pass the exact sign-test criterion ($12$/$4$ discordant, sign-test $p=0.077$). The cell is therefore a fragile positive rather than a robust significant result. The other five cells have bootstrap CIs that include zero, with the Belebele/gpt-4o lower bound touching zero. Figure~\ref{fig:forest_tt_back} visualizes the same numbers.

\subsection{Cross-Cell Pattern and Parser Sensitivity}

All non-zero point estimates have the same sign (five positive, one exactly zero); the cells share pipeline, items, and scoring and are not independent. The pattern is descriptive directional evidence; the item-clustered bootstrap of Section~\ref{sec:item_clustered} is the inferential summary of record.\footnote{Treating the five non-zero cells as independent draws under a no-effect null gives an exact sign-test $p = 2^{-5} = 0.0313$ one-sided. The independence assumption is not supported by the design and the binomial-tail interpretation does not apply.} The largest single cell estimate ($+0.047$ on MMMLU/gpt-5.4-mini) comes from the model with lowest parser validity ($0.825$ on $q_{\text{ZH}}$); excluding gpt-5.4-mini leaves three positive cells and one zero. The qualitative finding survives this exclusion; the $\geq 0.95$ parser-validity subset reduces to gpt-4o and gpt-4o-mini cells.

\subsection{Item-Clustered Bootstrap (Primary Inference)}
\label{sec:item_clustered}

Because the six cells share items, scoring, and pipeline, a cross-cell binomial test overstates evidence by treating correlated observations as independent. We therefore report an item-level cluster bootstrap as the primary inferential object: each cluster is one $(\text{benchmark}, \text{item})$ pair containing up to three Group-A model observations of $\text{acc}_{\text{EN}} - \text{acc}_{\text{back}}$. Resampling $n_{\text{clusters}}=296$ clusters with replacement ($B = 10{,}000$, seed $42$) gives \[ \widehat{\text{TT}}_{\text{cluster}} = +0.022, \quad 95\% \text{ CI} = [+0.001, +0.044]. \] The CI excludes zero, providing the paper's strongest single piece of evidence for a small positive translation tax averaged across items and Group-A models, after item correlation is accounted for. Two sensitivity checks: excluding gpt-5.4-mini (the parser-fragile model) gives $+0.015 \, [-0.008, +0.039]$, with the CI now including zero; restricting to MMMLU items only gives $+0.026 \, [-0.002, +0.054]$, with the lower bound just below zero. The item-clustered estimate is consistent with the cell-level pattern but is more conservative than a cross-cell sign test because it does not assume independence across cells; we treat it, rather than the binomial, as the primary quantitative summary.

\subsection{Group-Level Accuracy on $q_{\text{ZH}}$ (MMMLU)}

We tested whether English-centric models (Group A) outperform Chinese-optimized (Group B) or open-source multilingual (Group C) on translated Chinese MMMLU. Group means: $\mathrm{A_{en}} = 0.705$, $\mathrm{B_{zh}} = 0.694$, $\mathrm{C_{open}} = 0.750$. Group A and Group B are within $0.011$ of each other; Group C (with qwen2.5-72b at $0.807$) leads. \textbf{This does not support} the hypothesis that English-centric models systematically exploit cues to outperform Chinese-optimized models on the translated benchmark. The cue exploitation hypothesis, if it holds, must operate at item-level variation rather than aggregate accuracy.

\subsection{INCLUDE Non-Translation Control (E3)}

INCLUDE Chinese (545 four-choice items) was scored on six models: three Group-A and three Group-B. Aggregate accuracies (Wilson 95\% CI) and bootstrap gap MMMLU $-$ INCLUDE:
\begin{itemize}
\item \textbf{Group A (English-centric):}
gpt-4o $0.784$ vs $0.782$, gap $+0.002$ $[-0.064, +0.065]$;
gpt-4o-mini $0.674$ vs $0.645$, gap $+0.029$ $[-0.043, +0.101]$;
gpt-5.4-mini $0.803$ vs $0.791$, gap $+0.012$ $[-0.057, +0.076]$.
All three CIs include zero.
\item \textbf{Group B (Chinese-optimized):}
deepseek-chat $0.767$ vs $0.845$, gap $\mathbf{-0.078}$ $\mathbf{[-0.143, -0.015]}$;
qwen-max $0.727$ vs $0.851$, gap $\mathbf{-0.125}$ $\mathbf{[-0.190, -0.059]}$;
glm-4.5 $0.612$ vs $0.572$, gap $+0.040$ $[-0.037, +0.115]$.
Two of three Group-B CIs exclude zero in the negative direction; glm-4.5's CI
includes zero and is positive in sign.
\end{itemize}

\paragraph{Subgroup pattern.} Two of three Chinese-optimized models (deepseek-chat
$-7.8$pp; qwen-max $-12.5$pp) score substantially lower on translated MMMLU Chinese than on natively-sourced INCLUDE Chinese, with CIs excluding zero. The third Group-B model, glm-4.5, runs the other way ($+4.0$pp, CI including zero) and has substantially lower overall accuracy, making the gap estimate noisier. The two-of-three pattern is compatible with a subgroup reading in which stronger native Chinese training advantages models on native Chinese text, but is not a universal Group-B effect. INCLUDE and MMMLU differ in domain, register, and construction, so a Group-B model's higher INCLUDE score could also reflect distributional differences rather than translation-cue asymmetry; full identification would require item-matched native-vs-translated pairs that INCLUDE does not provide.

Group A models show small positive gaps consistent with zero; the Group A E3 estimates ($+0.002$, $+0.029$, $+0.012$) are similar in scale to E1 TT$_{\text{back}}$ estimates ($+0.029$, $+0.005$, $+0.047$).

\paragraph{Three readings of the Group A near-zero pattern.} (a) Cue exploitation is
small or absent for frontier English-centric models on this language pair; (b) MMMLU (academic) and INCLUDE (regional-exam) content distributions differ enough that aggregate comparison is uninformative; (c) the two benchmarks coincide in difficulty for these models. The Group-B pattern suggests a possible model-family interaction, but it does not rule out group-specific domain fit to INCLUDE-style regional exam items: Chinese-optimized models may be more familiar with Chinese regional-exam content and register, independent of any translation-cue effect on MMMLU. The six-model result is exploratory; three models (all Group C) remain unscored in this report.

\subsection{Annotation Calibration (E2)}
\label{sec:annotation_pilot}

A 30-item annotation calibration was run with one LLM-rubric annotator using a structured rubric covering three dimensions on a 5-point Likert scale. 26 items received complete annotations (4 fell on QC-excluded back-translations). Annotation distribution: cue score mean $3.08$ (sd $0.55$), cultural residue mean $2.58$ (sd $1.18$, range 1--5 with high\_school\_us\_history correctly identified at $5$ and abstract\_algebra at $1$), syntactic residue mean $2.58$ (sd $0.57$).

\paragraph{Calibration result.} Spearman correlations between cue score and item-level
TT signal: gpt-4o $\rho = -0.34$ ($p = 0.072$); gpt-4o-mini $\rho = 0.00$ (all 26 paired items tied); gpt-5.4-mini $\rho = +0.05$ ($p = 0.82$). The gpt-4o coefficient runs opposite the cue-exploitation hypothesis. Two factors block interpretation: of 26 annotated items, 25 produce TT signal $= 0$ (only one discordant pair, $96\%$ ties); the calibration uses a single LLM-rubric annotator, not two bilingual humans. The calibration is reported as a null result on rubric-based item-level cue exploitation under this protocol; it is not informative about the bilingual-human estimator at scale.

\subsection{Parser Validity Audit}

We report parser validity per model-set as a check on whether extraction failures bias accuracy estimates. 8 of 9 models on MMMLU achieve validity $\geq 0.978$; the exception is gpt-5.4-mini at $0.825$ on $q_{\text{ZH}}$, $0.864$ on $q_{\text{EN}}$, $0.870$ on $q^{\text{back}}$. Inspection of parser failures reveals empty raw responses, consistent with token-budget truncation on an extended-reasoning model. We report gpt-5.4-mini results on validity-restricted subsets; the strict-QC analysis in Section~\ref{sec:results} already excludes parser-failed items.

\section{Matched Naturalization Counterfactual (E4)}
\label{sec:e4_naturalization}

The three estimators in Section~\ref{sec:results} are proxies: they do not vary item content between conditions. This section reports a fourth estimator, E4, that holds item content fixed and varies only surface naturalness through a paired counterfactual rewrite of each translated Chinese item.

\paragraph{LLM-augmented protocol (disclosure).} The naturalization rewrites
were produced by a large language model (deepseek-chat) under a structured prompt that requires preservation of meaning, answer key, choice ordering, difficulty, and technical terminology, and a second large language model (gpt-4o) was used as a verifier scoring six dimensions per rewrite (semantic preservation, difficulty preservation, answer-key preservation, option-order preservation, cue removal, excessive-rewrite risk). This is an \emph{LLM-naturalization stress test}, not a bilingual-human naturalization counterfactual; reviewers should read the E4 treatment as ``LLM-rewritten Chinese with LLM verifier QC,'' not ``human-naturalized text.'' The prompts, raw rewrites, and verifier scores are released in the analysis package.

\paragraph{Sample and quality.} A 120-item stratified sample was drawn from
the 228-item MMMLU pilot: 30 high-residue items (Group A all correct on $q_{\text{EN}}$ but at least one wrong on $q_{\text{ZH}}$), 60 low-residue items (Group A all correct on both versions), and 21 disagreement items (Group A and Group B majority disagree). 54 distinct subjects, capped at four items per subject. One item failed the naturalization API; $n=119$ retained. LLM-verifier strict-QC pass rate was $119/119$ (mean semantic preservation $4.98$, mean difficulty preservation $5.00$, answer-key and option-order preserved on all items, mean cue removal $4.97$, mean excessive-rewrite risk $1.06$). These are LLM-verifier scores, not human QC.

\paragraph{Human QC and human two-coder validation ($n=50$).}
A bilingual blind QC was performed by the paper author on a stratified 50-item subset (13 high-residue, 9 disagreement, 28 low-residue), blind to per-model scoring records. Strict pass requires semantic and difficulty $\geq4$, answer-key and option-order pass, and excessive-rewrite risk $\leq2$. The author strict-pass rate is $49/50$; the only failure (\texttt{mmmlu\_zh\_00924}) dropped the I/II/III statement list, making the item unanswerable. On the author strict-pass subset, pooled $\Delta$ is $+0.022\,[-0.030,+0.078]$ excluding glm-4.5, with high-residue $\Delta=+0.133$ and low-residue $\Delta=-0.022$. We also include a second human QC of the same sheet by an independent human coder. This is human two-coder validation. The second human coder is stricter: $46/50$ strict pass, strict-pass agreement $94\%$ with the first author QC, and binary $\kappa=0.38$ (low because pass/fail is highly imbalanced). It flags three additional semantic or rewrite-risk concerns (state/nation, half-sister/ step-sister, and doctrinal-expansion cueing). On the intersection strict-pass subset ($n=46$), pooled $\Delta$ excluding glm-4.5 weakens to $+0.009\,[-0.042,+0.068]$, while the high-residue stratum remains directionally positive ($+0.125\,[0.000,+0.281]$) and low-residue remains near zero ($-0.023\,[-0.062,+0.008]$). The released supplement contains the first-coder QC sheet, the second-coder QC sheet, and the intersection sensitivity in \texttt{e4\_naturalization/human\_qc/} and \texttt{e4\_naturalization/second\_coder\_qc/}.

\paragraph{Scoring protocol.} Six models were scored: three Group A (gpt-4o,
gpt-4o-mini, gpt-5.4-mini) and three Group B (deepseek-chat, qwen-max, glm-4.5). A prompt-construction audit found that an earlier naturalized- condition scoring run duplicated choice labels: the naturalized choices retained an in-text leading ``A./B./C./D.'' label from the rewriter while the prompt builder also prepended canonical labels, producing options of the form ``A. A. text''. This affected the naturalized condition only; the translated condition uses the original answer-only prompt. The corrected protocol strips any leading $[A$--$D]$ label from each naturalized choice before prompt construction and uses the same prompt builder as the translated condition. All E4 numbers reported below use the corrected scoring protocol (release identifier \texttt{scoring\_v2/}). All translated-condition scores are unchanged. Parser-validity rates after correction: gpt-4o, gpt-4o-mini, qwen-max $1.00$; deepseek-chat $0.99$; gpt-5.4-mini $0.91$; glm-4.5 $0.34$. The naturalized prompt removes a format crutch glm-4.5 had relied on; the model is treated as a parser-failure outlier and excluded from the main analysis, included only in sensitivity. Per-item paired analysis below uses cells where both translated and naturalized scoring produced a valid letter.

\begin{figure}[h]
\centering
\includegraphics[width=\linewidth]{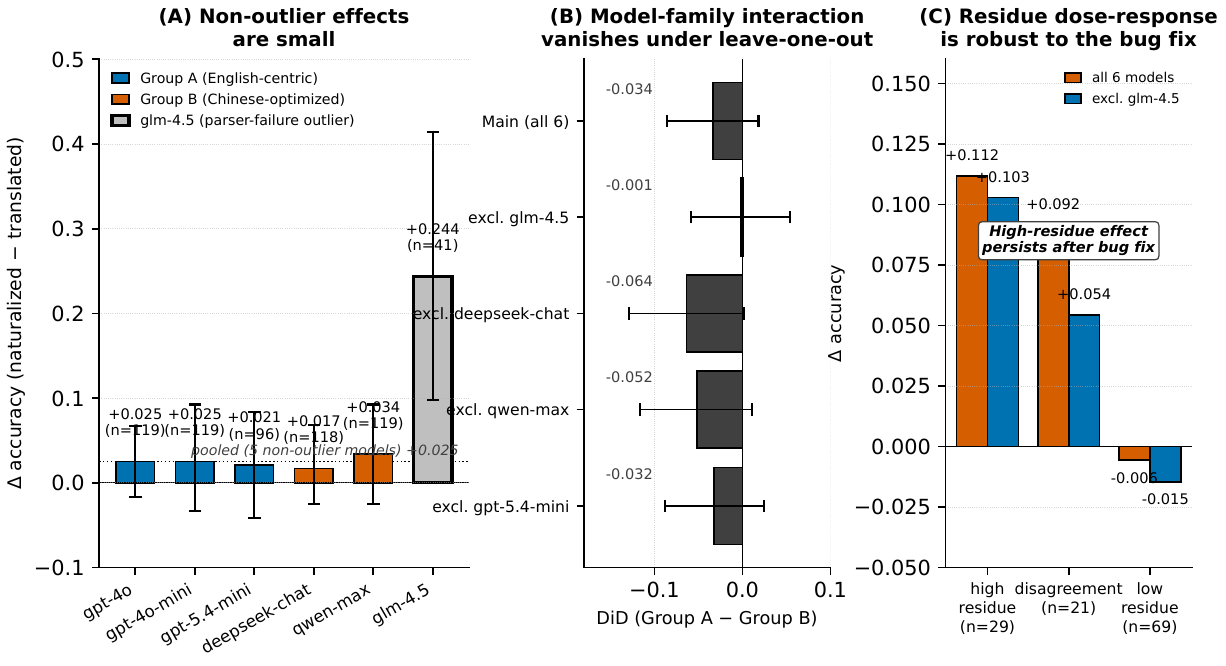}
\caption{After correcting the prompt-construction bug, E4 no longer supports
a stable model-family interaction. The remaining signal is item-level:
high-residue items benefit from naturalization while low-residue items do
not; glm-4.5 is treated as a parser-failure outlier.}
\label{fig:e4_killer}
\end{figure}

\paragraph{Item-clustered primary inference (E4 v2, main: 5 models, glm-4.5
sensitivity only).} Item-clustered bootstrap with $B=2{,}000$:
\begin{itemize}\itemsep1pt\parskip1pt
\item Pooled (5 non-outlier models): $\Delta = +0.025\,[-0.017,+0.071]$;
Group A: $+0.024\,[-0.009,+0.059]$;
Group B excl.\ glm-4.5: $+0.025\,[-0.017,+0.071]$;
DiD (A $-$ B excl.\ glm-4.5): $-0.001\,[-0.053,+0.058]$.
\item Pooled (all 6, glm-4.5 included): $\Delta = +0.039\,[+0.010,+0.070]$;
Group B (incl.\ glm-4.5 on its 41-item parser-valid subset):
$+0.058\,[+0.014,+0.104]$;
DiD: $-0.034\,[-0.087,+0.023]$ (CI crosses zero).
\end{itemize}
The model-family interaction reported in earlier drafts was driven by glm-4.5 in combination with the bug-affected prompt format. After the fix, E4 returns no statistically significant model-family interaction; the interaction observed in v1 is reclassified as an artifact of the combination of (i) double-labeled options and (ii) glm-4.5's degraded parser behaviour without that format crutch.

\paragraph{Stratum and leave-one-out sensitivity (Figure~\ref{fig:e4_killer}
B--C).} The dose-response pattern survives the bug fix: high-residue $\Delta = +0.103$ versus $-0.015$ on low-residue items (excl.\ glm-4.5); $+0.112$ versus $-0.006$ with glm-4.5 included. The high-to-low ratio is roughly $7\times$ in the main analysis (compared to the artefactual $\sim$$90\times$ in v1). Pooled $\Delta$ in leave-one-out configurations ranges $[+0.025, +0.045]$; the only excluded model that materially shifts the pooled estimate is glm-4.5. Full sensitivity tables are in \texttt{regression\_results\_v2.json}.

\paragraph{Interpretation and triangulation.} When item content is held fixed
correctly, E4 delivers a positive point estimate on high-residue items and a near-zero effect on low-residue items; the model-family interaction in v1 was a prompt-format artefact. The stricter second human QC weakens pooled E4 estimates, so E4 is treated as a diagnostic stress test rather than confirmatory evidence. The four estimators thus converge on one finding---residue-sensitive items are more fragile---and disagree on magnitude: E1 and E4 give small positive contrasts, while E2/E3 give null or model-family-internal patterns. No single estimator identifies a Translation Tax; the four together describe estimator- and item-dependent validity risks rather than a scalar correction. The supplement (\url{https://github.com/chi-mi-rvard/translation-tax-supplement}) recomputes all estimates including the corrected E4 protocol and QC sensitivities.

\paragraph{Limitations.} E4 uses an LLM rewriter and LLM verifier; the human
QC subset uses two human coders. The second human QC provides human two-coder validation. A human-naturalized subset (translated $q_{\text{ZH}}$ vs.\ human-rewritten $q_{\text{ZH}}$ on the same items) is not included in this submission. E2 is a 30-item single-annotator calibration, not a full bilingual-human estimator. glm-4.5's 0.34 parser-validity rate under the corrected naturalized prompt is reported as prompt sensitivity in that model rather than as a feature of translation residue. The raw outputs, bug-fix re-scoring, human QC, second human QC, and strict-subset sensitivities are released.

\bibliographystyle{plainnat}

\begin{thebibliography}{99}

\bibitem[Artetxe et al.(2020)]{artetxe2020translation}
Mikel Artetxe, Gorka Labaka, and Eneko Agirre.
\newblock Translation artifacts in cross-lingual transfer learning.
\newblock In \emph{EMNLP}, 2020.

\bibitem[Bandarkar et al.(2024)]{bandarkar2024belebele}
Lucas Bandarkar, Davis Liang, Benjamin Muller, Mikel Artetxe, Satya Narayan
  Shukla, Donald Husa, Naman Goyal, Abhinandan Krishnan, Luke Zettlemoyer, and
  Madian Khabsa.
\newblock The {B}elebele benchmark: a parallel reading comprehension dataset in
  122 language variants.
\newblock In \emph{ACL}, 2024.

\bibitem[Clark et al.(2020)]{clark2020tydiqa}
Jonathan~H. Clark, Eunsol Choi, Michael Collins, Dan Garrette, Tom Kwiatkowski,
  Vitaly Nikolaev, and Jennimaria Palomaki.
\newblock {T}y{D}i {QA}: A benchmark for information-seeking question answering
  in typologically diverse languages.
\newblock \emph{TACL}, 8:454--470, 2020.

\bibitem[Conneau et al.(2018)]{conneau2018xnli}
Alexis Conneau, Ruty Rinott, Guillaume Lample, Adina Williams, Samuel Bowman,
  Holger Schwenk, and Veselin Stoyanov.
\newblock {XNLI}: Evaluating cross-lingual sentence representations.
\newblock In \emph{EMNLP}, 2018.

\bibitem[Hendrycks et al.(2021)]{hendrycks2021mmlu}
Dan Hendrycks, Collin Burns, Steven Basart, Andy Zou, Mantas Mazeika, Dawn
  Song, and Jacob Steinhardt.
\newblock Measuring massive multitask language understanding.
\newblock In \emph{ICLR}, 2021.

\bibitem[Mizrahi et al.(2024)]{mizrahi2024multiprompt}
Moran Mizrahi, Guy Kaplan, Dan Malkin, Rotem Dror, Dafna Shahaf, and Gabriel
  Stanovsky.
\newblock State of what art? a call for multi-prompt {LLM} evaluation.
\newblock \emph{TACL}, 12:933--949, 2024.

\bibitem[Ponti et al.(2020)]{ponti2020xcopa}
Edoardo~Maria Ponti, Goran Glava\v{s}, Olga Majewska, Qianchu Liu, Ivan
  Vuli\'{c}, and Anna Korhonen.
\newblock {XCOPA}: A multilingual dataset for causal commonsense reasoning.
\newblock In \emph{EMNLP}, 2020.

\bibitem[Romanou et al.(2025)]{romanou2025include}
Angelika Romanou, Negar Foroutan, Anna Sotnikova, Sree~Harsha Tanneru, Zeming
  Chen, Antoine Bosselut, and Syrielle Montariol.
\newblock {INCLUDE}: Evaluating multilingual language understanding with
  regional knowledge.
\newblock In \emph{ICLR}, 2025.

\bibitem[Sitaram(2025)]{sitaram2025}
Sunayana Sitaram.
\newblock Coverage, representativeness, trust and scientific rigor in
  multilingual benchmark evaluation.
\newblock Invited talk, NeurIPS 2025 Workshop on Centering Low Resource
  Languages and Cultures.

\bibitem[Wu et al.(2025)]{wu2025}
Minghao Wu, Weixuan Wang, Sinuo Liu, Huifeng Yin, Xintong Wang, Yu Zhao, Chenyang Lyu, Longyue Wang, Weihua Luo, and Kaifu Zhang.
\newblock The bitter lesson learned from 2{,}000+ multilingual benchmarks.
\newblock arXiv:2504.15521, 2025.

\bibitem[Zhuo et al.(2024)]{zhuo2024prosa}
Jingming Zhuo, Songyang Zhang, Xinyu Fang, Haodong Duan, Dahua Lin, and Kai
  Chen.
\newblock {ProSA}: Assessing and understanding the prompt sensitivity of
  {LLM}s.
\newblock In \emph{EMNLP Findings}, 2024.

\end{thebibliography}


\end{document}